\documentclass[10pt, twocolumn]{article}
\usepackage{cvpr}
\usepackage[most]{tcolorbox} 
\usepackage{lipsum}
\usepackage{times}
\usepackage{epsfig}
\usepackage{wrapfig, blindtext}
\usepackage{graphicx}
\usepackage{amsmath}
\usepackage{cite}
\usepackage{booktabs}
\usepackage{amssymb}
\usepackage{fancyhdr}
\usepackage{parskip}
\usepackage{lettrine}
\usepackage{tikz}
\usetikzlibrary{arrows}
\usetikzlibrary{arrows.meta}
\usepackage{color}
\usepackage{xifthen}

\usepackage[breaklinks=true,bookmarks=false]{hyperref}
\usepackage{fancyhdr}
\cvprfinalcopy 
\def\cvprPaperID{****} 
\begin{document}

\title{Beyond Attention: Toward Machines with Intrinsic Higher Mental States}

\author{Ahsan Adeel$^{}$\thanks{Conscious Multisensory Integration (CMI) Lab, University of Stirling, Stirling, UK and Computational Neuroscience Laboratory, John Radcliffe Hospital, Oxford, UK.\\
Email: ahsan.adeel1@stir.ac.uk}}

\newcommand*\samethanks[1][\value{footnote}]{\footnotemark[#1]}

\maketitle

\begin{abstract} 
Attending to what is relevant is fundamental to both the mammalian brain and modern machine learning models such as Transformers. Yet, determining relevance remains a core challenge, traditionally offloaded to learning algorithms like backpropagation. Inspired by recent cellular neurobiological evidence linking neocortical pyramidal cells to distinct mental states, this work shows how models (e.g., Transformers) can emulate high-level perceptual processing and awake thought (imagination) states to pre-select relevant information before applying attention. Triadic neuronal-level modulation loops among questions ($Q$), clues (keys, $K$), and hypotheses (values, $V$) enable diverse, deep, parallel reasoning chains at the representation level and allow a rapid shift from initial biases to refined understanding. This leads to orders-of-magnitude faster learning with significantly reduced computational demand (e.g., fewer heads, layers, and tokens), at an approximate cost of $\mathcal{O}(N)$, where \textit{N} is the number of input tokens. Results span reinforcement learning (e.g., CarRacing in a high-dimensional visual setup), computer vision, and natural language question answering.
\end{abstract}
\section{Introduction}
Beyond the attention mechanisms used in modern artificial neural networks, such as the standard Transformer \cite{vaswani2017attention} and its variants, Perceiver \cite{ jaegle2021perceiver} and Flamingo \cite{alayrac2022flamingo}, recent cellular neurobiological evidence \cite{Phillips2024cellular} strongly suggests that transitions between distinct mental states (from wakefulness to slow-wave (SW) sleep and rapid eye movement (REM) sleep) modulate momentary interactions in layer 5 pyramidal two-point neurons (TPNs; Figure 1) \cite{larkum1999new, phillips2023cooperative, larkum2013cellular, larkum2022dendrites, SchumanAnnual, poirazi2020, larkum2018perspective, shine2016dynamics, shine2019human, shine2019neuromodulatory, shine2021computational, schulz2021gaba}. These interactions occur between inputs from the external world (receptive field, RF\footnote{RF: External world. The area in the sensory periphery where stimuli can affect the electrical activity of sensory cells.}) and the internal world (contextual field, CF\footnote{CF: Internal world. Information from a wide range of cortical and subcortical sources, including feedback.}), converging at two functionally distinct integration sites: (i) a basal site that integrates input to the soma and basal dendrites, and (ii) an apical site, integrating input to the apical dendrites.
\begin{tcolorbox}[colback=white, colframe=black, boxrule=0.5pt, sharp corners]
\noindent
\textbf{Significance Statement and Future Directions: } In addition to demonstrating fast and economical learning by emulating high-level perceptual processing and awake (imaginative) thought states, the investigation presented here suggests that the mind constructs relevant context from available information to justify initial thoughts. If an initial thought is misleading, reaching a correct conclusion may require significantly more time, or may never occur at all, due to constrained cognitive resources. This leads to the following hypothesis: profound awareness arises from the ability to simultaneously evolve latent, rich, and coherent Q, K, and V representations in lockstep with reality, that is, at a pace matching the rate at which reality itself changes, as judgments are formed. Without this alignment, perception may be misleading or nothing more than an illusion. This applies only to observable reality.
\end{tcolorbox}

\begin{figure} 
	\centering
	\includegraphics[trim=0cm 0cm 0cm 0cm, clip=true, width=0.5\textwidth]{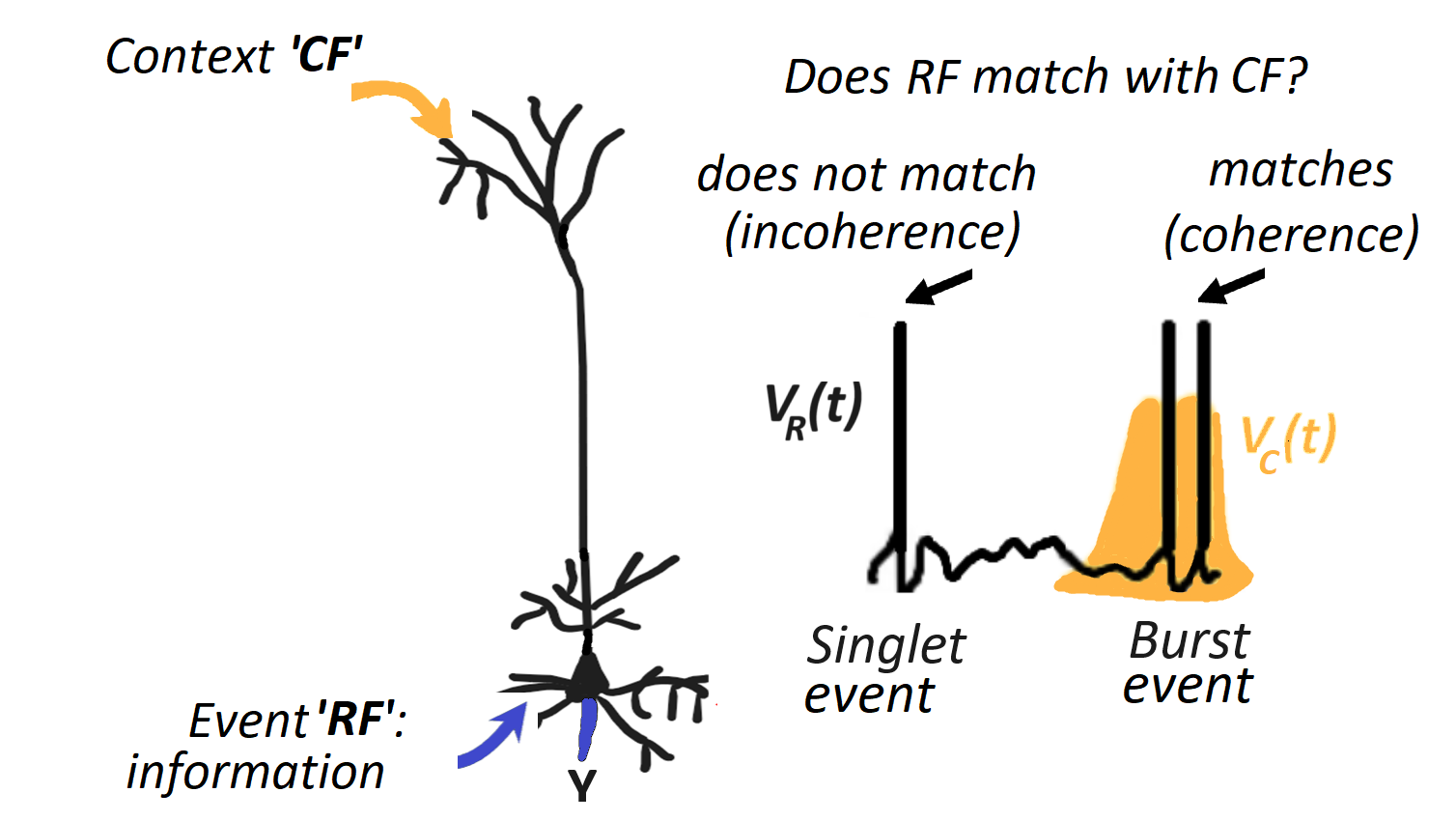}
	\caption{A pyramidal two-point neuron in the mammalian neocortex \cite{larkum1999new, Phillips2024cellular} integrates information at two functionally distinct sites. Its apical dendrites integrate CF at the contextual integration zone, while RF is integrated at the basal compartment, also known as the feedforward integration zone. The neuron triggers high-frequency firing (bursting) when RF and CF are matched in time, that is, when both the basal and apical zones are depolarized. This results in the amplification of coherent signals, enabling enhanced processing of contextually relevant information.}
	\label{l5pc}
 \vspace{-1em}
\end{figure}

During sensory perception, apical dendrites (CF), when receiving moderate to high input, typically amplify the transmission of information conveyed by feedforward (FF) basal or perisomatic input (RF). This context-sensitive cooperative amplification supports cognitive abilities excluding internally generated imagery and thought.\\
When the apical input is high to maximal, it can independently drive the axonal spiking output, a condition associated with REM sleep and internally generated thought during wakefulness (imagination). When the apical input has no effect on the neuron’s current output, this state is linked to SW sleep. The flexible interaction between RF and CF inputs is suggested to be the hallmark of conscious processing \cite{aru2020cellular, storm2024integrative, marvan2021apical}. Dysfunctional interactions between RF and CF inputs have been linked to intellectual learning disabilities \cite{nelson2021dendritic, granato2024dysfunctions}.\\
Inspired by these cellular mechanisms, a Cooperative Context-sensitive Cognitive Computation mechanism, abbreviated as $Co^4$, has been developed for Transformers. Drawing on high-level perceptual processing and wakeful (imaginative) thought states, it enables on-the-fly selection of relevant information before attention is applied, through triadic modulation loops among latent questions (Q), clues (keys, K), and hypotheses (values, V), allowing for a rapid transition from raw input to refined conclusions.\\
In the high-level perceptual processing state, K neurons provide moderate-to-high contextual input to Q neurons, amplifying only those Q elements that align with K and evolving a list of questions based on this context. Similarly, Q neurons provide contextual feedback to K neurons, amplifying only those K elements coherent with Q, thereby refining the context representation. In the wakeful (imaginative) state, hypotheses (V) evolve based on both Q and K, which together provide a high-to-maximal level of contextual input, followed by a concluding attention operation.\\
In this process, Q dynamically adapts its inquiries based on the evolving K and V, while K adjusts its representation in response to posed questions and the evolving V, with V emerging from their bidirectional interaction. As a result, the system rapidly identifies coherent Q–K–V relationships, which are then processed by a lightweight attention mechanism across diverse, deep, and parallel reasoning chains.\\
Unlike standard Transformers, which rely on Q–K similarity without internal reasoning or feedback and require deep stacks with costly $\mathcal{O}(N^2)$ attention, this biologically inspired alternative enables efficient, context-sensitive reasoning at a representation-level Chain-of-Thought, with a significantly lower computational cost of approximately $\mathcal{O}(N)$. This approach significantly accelerates learning and reduces computational demand compared to standard Transformers.
\section{Background}
The mechanism of flexibly combining top-down (CF) and bottom-up (RF) information streams provides the cellular foundation for computational theories of neocortical functioning \cite{aru2020apical, bachmann2020dendritic, marvan2021apical, aru2020cellular}. One of the most prominent computational frameworks for a dual-stream information flow is based on the principle of free energy minimization \cite{friston2010free} or predictive coding (PC) \cite{spratling2017review, bastos2012canonical, friston2005theory} \cite{friston2018deep, friston2017graphical}. \\
In PC, predictions as CF sent from higher to lower levels of processing attempt to attenuate the sensory signals RF that align with those predictions. Unaligned predictions are passed on to higher levels as prediction errors. Evidence on apical function contradicts the notion of subtractive prediction error coding because the two distinct sources lead to amplification when they agree (i.e., they are cooperative, not subtractive) \cite{marvan2024cellular}. This does not contradict the free energy principle, however, because that principle does not necessarily imply subtractive coding. \\
Computational neuroscientists have proposed various biologically plausible learning algorithms, including the recent burst-dependent synaptic plasticity (BDSP) \cite{payeur2021burst, Greedysingle}, which distinguishes between a single spike and a burst of spikes. It uses the top-down (apical) zone of the TPN to receive FB signals (bursting rate $b$) from higher perceptual levels as CF. The difference between the instantaneous and moving average of $b$ is multiplied by the event rate \textit{(e)} or RF to update weights \textit{(W)}. BDSP approximates loss-function gradients similar to backpropagation and performs comparably on complex image patterns, addressing the credit assignment problem effectively. However, BDSP and similar algorithms \cite{guerguiev2017towards, sacramento2018dendritic, IllingLocal, Greedysingle, zenke2017continual, kirkpatrick2017overcoming, kastellakis2016linking, bono2017modeling, limbacher2020emergence} focus mainly on apical inputs for learning. While their learning is TPN-inspired, their processing is not.\\
In contrast, there is ample neurobiological evidence suggesting that context serves as a modulatory factor \cite{marvan2024cellular, schulz2021gaba, kay2020contextual, kay2022comparison, phillips2023cooperative}. Neurophysiologists have proposed several biologically plausible asynchronous modulatory (MOD) transfer functions \cite{schulz2021gaba, kay2020contextual, kay2022comparison, phillips2023cooperative} and their latest iterations \cite{pastorelli2023two, graham2024transfer}; however, whether there are many applications in which deep neural nets (DNNs) inspired by these transfer functions can outperform Point Neurons (PNs) \cite{adeel2022unlocking}-driven DNNs is yet to be seen.\\
Recently a new TPN MOD function \cite{adeel2022unlocking, muckli_2023_8380094, raza2024overlooked} incorporated not only FB from higher perceptual levels but also simultaneous events across hierarchies while processing FF information \cite{aru2020cellular, bachmann2020dendritic, shin2021memories, adeel2022unlocking, muckli_2023_8380094}. This MOD function uses CF to split the RF into coherent and incoherent streams at the cellular level, recombining only the coherent ones. The MOD function assigns greater weight to CF, amplifying or attenuating RF based on CF's strength. \\Recent computational results with convolutional neural nets (CNNs) \cite{adeel2022unlocking} have shown that TPNs-inspired CNN with this MOD function minimizes the transmission of large amounts of conflicting FF information to higher perceptual levels, greatly reducing (by orders of magnitude) the number of neurons required to process large amounts of heterogeneous real-world audio-visual data compared to PN-inspired CNNs \cite{adeel2022unlocking}. The most recent study in cellular psychology \cite{Phillips2024cellular} links these types of MOD functions to apical drive (AD), which is utilized in $Co^4$.
\section{Varying Strengths of RF and CF Across Distinct Mental States: An Illustrative Example}
Before introducing the $Co^4$ architecture, this section illustrates how the strength of RF and CF inputs, and their interaction, varies across high-level perceptual processing and awake thought (imagination) states.\\
Read the ambiguous text in Figure 2 quickly at first, and then slowly \cite{kahneman2011thinking, selfridge1955pattern, rumelhart1986parallel}. When read quickly, our perception operates in a fast mode: automatic, pre-reflective, and anticipatory, neither too abstract nor too concrete \cite{parnas2021double, parnas2024phenomenological, blankenburg2001first}. We are able to focus on relevant information and decode it with a reasonable degree of confidence and coherence within that context. This corresponds to high-level perceptual processing \cite{Phillips2024cellular}: a state of being awake and conscious, characterized by basic, intuitive judgments sufficient for routine activities and everyday interactions, also known as common sense \cite{parnas2021double, parnas2024phenomenological, blankenburg2001first}. In this state, the strength and interaction of both RF and CF inputs range from moderate to high (Mod–High), mediated by cholinergic, noradrenergic, and orexinergic systems \cite{Phillips2024cellular}.\\
Now, if we slow down and take more time to interpret the information from different perspectives, we begin to perceive how identical characters can be interpreted differently depending on the context. For instance, consider the second character in both words in the first row, although visually identical, it is interpreted differently. Likewise, the first character in the second row and the second character in the third row are the same, yet each is perceived uniquely. This reflects a wakeful (imaginative) thought state \cite{Phillips2024cellular}: a heightened state of awareness and imagination that involves deep\footnote{Deep refers to the complex integration of diverse CFs at the cellular level, enabled by TPNs.}, slow, sequential, structured, deliberate, reflective, and well-justified judgments in context \cite{vlaev2018local}. In this case, the strength and interaction of RF and CF inputs range from high to maximal (High–Max), mediated by cholinergic, noradrenergic, and orexinergic pathways \cite{Phillips2024cellular}.
\\Overall, the combination of high-level perceptual processing and awake thought helps us navigate uncertainty and explore deeper, more nuanced, and less obvious meanings, moving beyond the literal interpretations offered by attention alone.
\section{$Co^4$ Architecture}
Going beyond the standard interaction of Q, K, and V in Transformers \cite{vaswani2017attention, jaegle2021perceiver, alayrac2022flamingo}, which collapses the rich functional distinctions between the RF and CF into an overly simplistic form, the $Co^4$ mechanism (Figure 3) dynamically tunes the latent question vectors ($Q_m$), i.e., \textit{what should be looked for}, based on evolving clues ($K_m$) and hypotheses ($V_m$). $K_m$ evolves based on both $Q_m$ and $V_m$, while $V_m$ evolves based on $K_m$ and $Q_m$. The triadic feedback loops allow latent $Q_m$ tokens to adapt dynamically to both available information and emerging hypotheses. The modulated keys ($K_m$) adapt each input token based on \textit{who's asking}, i.e., the latent $Q_m$, and their evolving hypothesis, while the modulated values ($V_m$) evolve specifically in response to both the latent questions and contextual clues, prior to the application of attention.
\begin{figure} 
	\centering
	\includegraphics[trim=0cm 0cm 0cm 0cm, clip=true, width=0.15\textwidth]{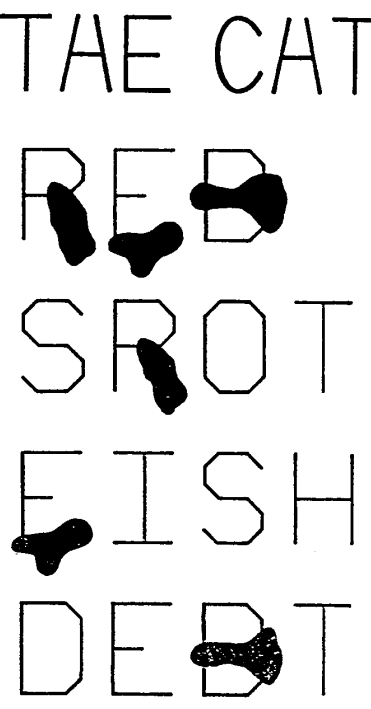}
	\caption{An example of ``thinking fast and slow" as discussed in \cite{kahneman2011thinking}, illustrates how solving a riddle can involve combining rapid, intuitive processing in high-level perceptual state (fast thinking) with more deliberate, reflective refinement in the awake thought state (slow thinking).} 
\vspace{-1.4em}
\end{figure}
\\In effect, the input token implicitly asks:  
\begin{quote}
\textit{“How should I present myself based on what this latent is trying to understand?”}
\end{quote}
The execution of this reciprocally adaptive loop is made possible through the cooperation of TPNs. For example, when \textit{L} latent $Q$ tokens are used, $Co^4$ generates \textit{L} separate sets of $(Q_m, K_m, V_m)$, each experiencing a custom triadic interaction, although after this interaction, they are averaged to avoid implementation complexity and have a simple attention map of $\mathcal{O}(L \times N)$, virtually each latent,  produces its own attention map over the input sequence. \\
This multi-perspective deep reasoning at the layer level helps explain why $Co^4$ achieves strong performance with significantly fewer layers and attention heads compared to standard Transformers. $Co^4$ incorporates a high-level symbolic (human-readable) representation similar to Symbolic AI, also known as Good Old-Fashioned AI (GOFAI). 
\begin{figure*} 
	\centering
	\includegraphics[trim=0cm 0cm 0cm 0cm, clip=true, width=1\textwidth]{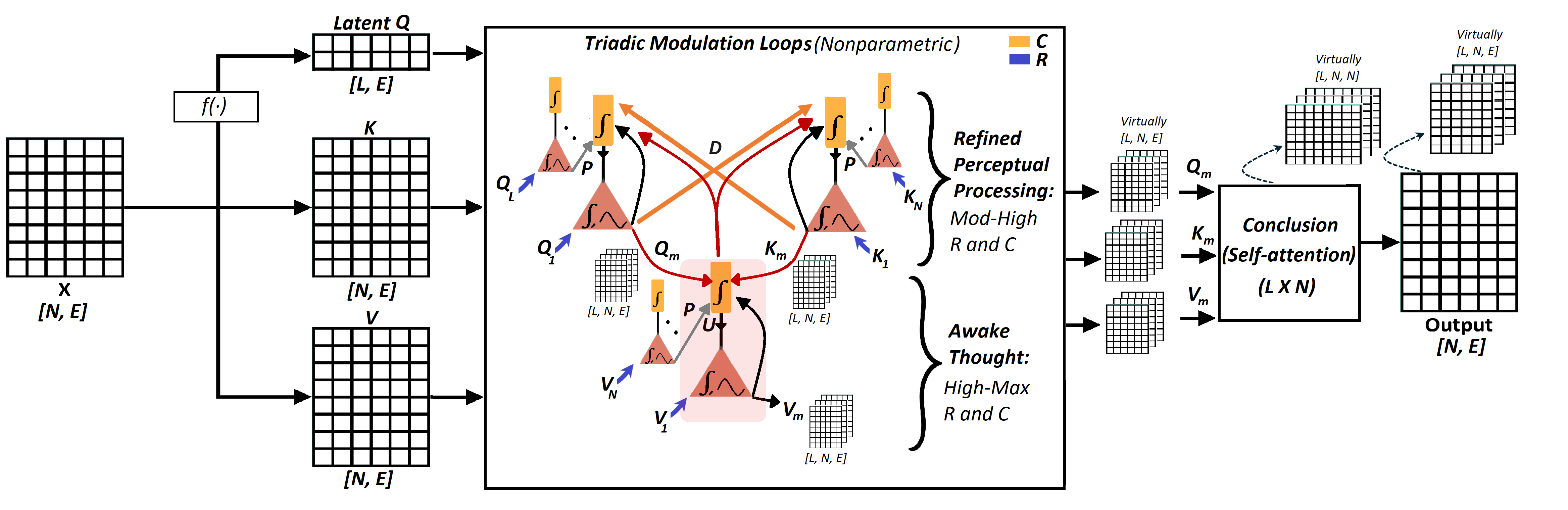}
	\caption{$Co^4$ architecture: $N$ denotes the number of input tokens, and each token has an embedding dimension of $E$. $Q_1$, $Q_2$,...,$Q_L$ represent the latent query tokens input to the associated Q-TPNs. $K_1$, $K_2$,...,$K_N$ represent the Key tokens input to the associated K-TPNs. $V_1$, $V_2$,...,$V_N$ represent the Value tokens input to the associated V-TPNs. This configuration forms part of the “seeing” state (i.e., sensory processing). In the “seeing as” state (i.e., perceptual and interpretive state), triadic modulation loops among questions ($Q$), clues (keys, $K$), and hypotheses (values, $V$) are executed through distal (D) and universal (U) contexts. Proximal (P) context represents normalization via information from neighboring neurons in the same population, including the prior information from the same neuron. The TPNs associated with $Q$, $K$, and $V$ are assumed to be analogous to three subtypes of pyramidal neurons, although their exact correspondence to neurobiologically distinguished subtypes is still under investigation. Through varying states of mind, high-level perceptual processing and wakeful thought, diverse, parallel reasoning chains are enabled. This mechanism incurs a computational cost of $\mathcal{O}(N \cdot L)$, where $L$ is a small fraction of the input length, making the overall cost approximately $\mathcal{O}(N)$. The triadic modulation loops, based on element-wise operations, add a nominal cost of $L \cdot N \cdot E$, which is significantly lower than that of the feedforward residual network used in standard Transformer blocks, a component $Co^4$ does not require. $Co^4$ can be viewed as a parallel, representation-level, silent yet deep form of Chain-of-Thought (CoT) reasoning \cite{wei2022chain} (a quiet mind), enabling multi-perspective inference without requiring sequential token-level generation, much like the brain’s cortico-thalamic modulation \cite{aru2020cellular, Phillips2024cellular, storm2024integrative}.}
	\label{l5pc}
 \vspace{-1em}
\end{figure*}
\begin{figure} 
	\centering
	\includegraphics[trim=0cm 0cm 0cm 0cm, clip=true, width=0.5\textwidth]{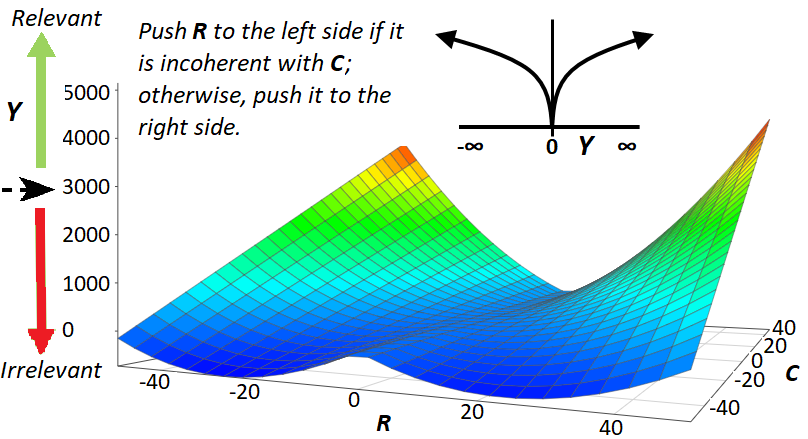}
	\caption{The TPN-inspired asynchronous MOD function (unnormalized) uses CF, denoted as $C$, to selectively amplify or attenuate the transmission of relevant and irrelevant RF, denoted as $R$, respectively. C serves as the driving force because the strong contextual signals from neighboring neurons can override even strong but noisy $R$ signals at the single-cell level. The combined perceptions of multiple neighboring neurons, each drawing from diverse and contextually relevant sources, carry more weight than the output of any individual neuron. This mechanism enhances cooperation between R and C for the task at hand. This behavior becomes particularly significant at higher levels of abstraction, where imagination begins to emerge and direct perception becomes less dominant. }
	\label{l5pc}
 \vspace{-1em}
\end{figure}
\begin{figure} 
	\centering
	\includegraphics[trim=0cm 0cm 0cm 0cm, clip=true, width=0.5\textwidth]{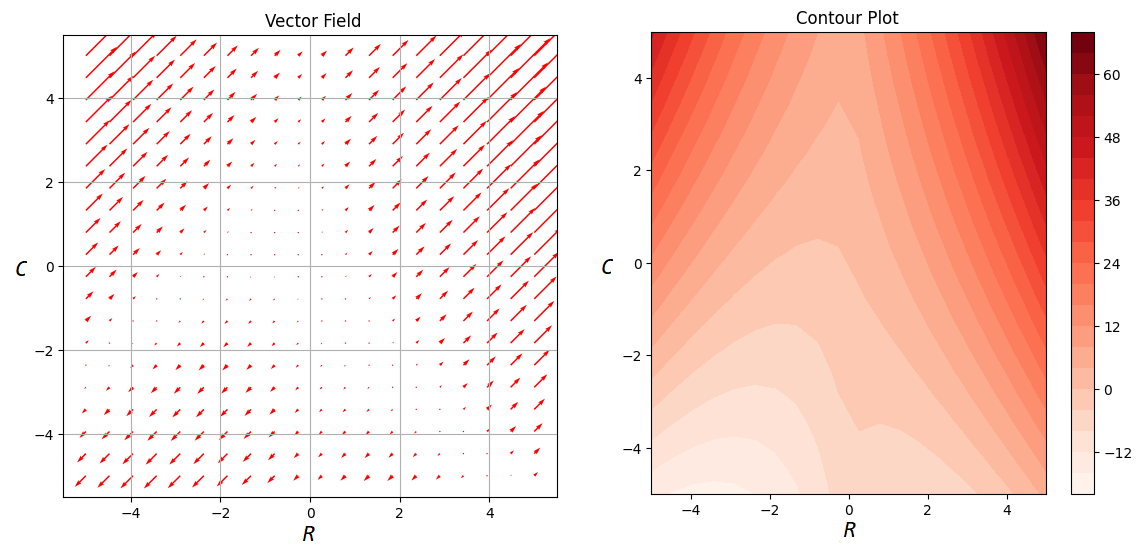}
	\caption{Vector field visualization and contour plots for deeper insight into the MOD function. The MOD function utilizes C to split R into coherent and incoherent R streams. The MOD function assigns greater weight to C; if C is high, and R is negative, the output of the function in eq (1) is positive (i.e. it makes  f(C,R) have a different sign from R). Conversely, if  C is low, R is diminished regardless of its own value.}
	\label{l5pc}
 \vspace{-1em}
\end{figure}
\vspace{0.5em}
\\Specifically, given an input of $N$ tokens with embedding size $E$, TPNs associated with latent $Q$, $K$, and $V$ extract:
\begin{enumerate}
    \item A set of modulated evolving questions ($Q_m$), dynamically tailored to the evolving context ($K_m$) and hypothesis ($V_m$).
    \item An evolving context ($K_m$) shaped in response to these questions ($Q_m$) and hypothesis ($V_m$).
    \item A set of possible hypotheses ($V_m$) generated from the coherent interaction of $Q_m$ and $K_m$.
    \item Each latent virtually generates its own attention map over the input sequence, even though this does not occur explicitly in implementation.
\end{enumerate}

To generate coherent $Q_m$, $Q$ must act as the RF, while $K$ and $V$ serve as the CF. To generate coherent $K_m$, $K$ must act as the RF, with $Q$ and $V$ as CF. In this state, the strength and interaction of RF and CF inputs range from moderate to high, representing the high-level perceptual processing state~\cite{Phillips2024cellular}. To generate coherent hypotheses $V_m$, $V$ must act as the RF, while both $Q$ and $K$ serve as CF. As we transition from fast to slow thinking (i.e., the wakeful state), the influence of CF on the interpretation of RF increases, implying that the interaction between RF and CF shifts from high to Max~\cite{Phillips2024cellular}.\\
Each TPN uses an asynchronous MOD function\footnote{There may be several other applicable variations of such functions, including those proposed elsewhere.} (Figure 4) to selectively amplify relevant RF transmissions and attenuate irrelevant ones within the given CF. This asynchronous MOD function, termed the \textit{Cooperation Equation}, is defined as:
\begin{equation}
\text{Cooperation}(R, C) = \text{ReLU6}\left(R^2 + 2R + C(1 + |R|)\right)
\end{equation}

where $R$ represents the RF and $C$ represents the CF. ReLU6 is a variant of the Rectified Linear Unit (ReLU) activation function commonly used in deep learning models and defined as:
\[
\mathrm{ReLU6}(x) \;=\;\min\!\bigl(6,\;\max(0,\,x)\bigr)
\]

In this equation, C as a `modulatory force' pushes the information to the positive side of the activation function (e.g., ReLU6) if R is relevant, otherwise to the negative side. In essence, strong C can discourage or encourage amplification of neural activity regardless of R's strength (strong or weak).  The value of C captures the strength of the signal, but different approximate ranges of this value are distinctly associated with different mental states. The differing behaviour reflects the nonlinearity in the interaction between C and R. \\
This behavior aligns with evidence showing that ``thinking" involves AD during wakefulness, particularly when both acetylcholine (ACh) and sodium (Na) ion levels are elevated \cite{Phillips2024cellular}. It is consistent with high-resolution neuroimaging studies \cite{muckli_2023_8380094, Phillips2024cellular} showing that imagination activates perception-related brain regions by generating missing information from memory, implying that CF can override RF. This phenomenon is also known as \textit{filling in}.  \\
While there is no independent parameter explicitly defining the importance or strength of C, its non-linear influence in the behavior of eq (1) allows its effect to vary dynamically across different processing stages. This non-linear amplification can be interpreted as capturing the importance of the context: when the context is strongly aligned or misaligned with the response, its effect is magnified. This behavior effectively implements the idea that the influence of context is stage-dependent, even without introducing a dedicated parameter for the “degree of importance”. \\
To provide a more precise and deep insight into the MOD function, the vector field visualization and contour plots are presented in Figure 5 showing how the output of the equation varies across a range of values for R and C. The MOD function uses C to conditionally segregate the coherent and incoherent R signals. C is the driving force; if C is high, R is amplified regardless of its own strength and if C is low, R is attenuated regardless of its own strength. \\ 
The vector field plot (left) shows the direction and magnitude of the output across the space. The contour plot (right) displays the magnitude of the output across the R-C space where the contour lines show levels of equal magnitude, giving us insight into how the equations behave differently across the space. \\
Specifically, the vector field plot illustrates the gradients (partial derivatives) of the MOD function. Each arrow shows the gradient's direction and magnitude at specific points, where the direction indicates the steepest ascent of the function, and the length reflects the rate of change. Diagonal arrows, especially in regions with positive R, suggest that both R and C contribute positively to the gradient. Dense or long arrows highlight areas of rapid change (steep slopes), while sparse or short arrows indicate slower changes, which filter and reshape representations—mitigating vanishing gradient by modulating abrupt transitions when context and signal are mismatched.\\
The contour lines, on the contour plot, connect points of equal function value, with closer lines indicating rapid changes and spaced lines indicating slower changes. The color gradient reflects function magnitude, with darker colors representing higher values. The color bar on the side provides a reference for these values. \\
Below are the alternative well-established TPNs-inspired asynchronous MOD functions \cite{kay1998contextually, phillips2023cooperative}. In these modulatory transfer functions ($T_{M}$) eq (2-5), R is the driving force i.e., if R is absent or strong, C has no role to play. 
\begin{equation}
T_{M1}(R, C) = \frac{1}{2}R(1+exp(RC))
\end{equation}
\begin{equation}
T_{M2}(R, C) = R+RC
\end{equation}
\begin{equation}
T_{M3}(R, C) = R(1+tanh(RC)) 
\end{equation}
\begin{equation}
T_{M4}(R, C) = R(2^{RC})
\end{equation}
The use of eq (1) was guided by empirical experiments with various nonlinear RF functions, including those presented above and their latest iterations \cite{graham2025context}, with the chosen version performing well. Nevertheless, further critical testing is needed, including testing Partial Information Decomposition \cite{graham2024transfer} resulting from eq (1).
\subsection{Computational Complexity}
$Co^4$ has a computational complexity of 
\[
\mathcal{O}(L \cdot N + \alpha)
\]
where \( N \) is the number of input tokens (patches or words), \( L \) is the number of latent tokens, and \( \alpha \) accounts for additional element-wise operations.  Instead of full attention between all \( N \) tokens,
\[
\Rightarrow \mathcal{O}(N^2),
\]
the model, similar to latent Transformers \cite{bi2024deepseek}, restricts this to \( N \times L \) interactions where \textit{L}
is a small fraction of the input length \textit{N},
\[
\Rightarrow \mathcal{O}(N \cdot L) \approx
\mathcal{O}(N)
\]
The element-wise operations in triadic modulation loops are significantly less expensive than matrix multiplications and lower than that of the FF residual network used in standard Transformer blocks, which $Co^4$ does not require. $Co^4$ scales linearly with input length, in contrast to the quadratic scaling in standard Transformers. For more explicit complexity analysis, the dominant MAC (multiply--accumulate) operations in standard Transformer model are approximated as:
\[
\textbf{MACs}_{\text{standard}} \approx L\Big(\,P\,E^2 + \,P^2\,E\Big)
\]
Where $E$, $P$, and $L$ represent the embedding dimension, number of tokens (or patches), and the number of layers, respectively. The term $\,P\,E^2$ represents the cost of the Q, K, V projections plus the FF network. The term $\,P^2\,E$ is due to the self--attention mechanism, which scales quadratically with $P$.\\
For the $Co^4$ model, the dominant MAC operations are approximated as:
\[
\textbf{MACs}_{\text{$Co^4$}} \approx L\Big(L_q\,E^2 + \,P\,E^2 + \,L_q\,P\,E\Big)
\]
$L_q$ represent the number of latent (query) tokens (with $L_q \ll P$).\\
The term $L_q\,E^2$ arises from projecting the $L_q$ latent queries. The term $\,P\,E^2$ corresponds to the key and value projections. The term $\,L_q\,P\,E$ reflects the cost of  triadic modulation loops (emulating high-level perceptual processing and wakeful thought states), which scales linearly with $P$ since $L_q$ is small. 
\\Overall, the standard Transformer includes a quadratic term ($\,P^2\,E$), while the $Co^4$ model reduces this to a linear term ($\,L_q\,P\,E$) by using a small set of latent queries.  For large $P$, the standard Transformer becomes computationally expensive due to the quadratic attention cost. The $Co^4$ model offers significant savings by leveraging a fixed, small number of latent queries, thereby reducing the computational complexity.
\section{Results}
The goal of this paper is to assess the power of the standard Transformer and $Co^4$ on a few quick-to-run tests using exactly the same architecture and hyperparameters. A more detailed analysis will be detailed in a follow-on paper. For 
reinforcement learning (RL) tasks, $Co^4$ performance is compared with \cite{tang2021sensory}. For CIFAR‑10, $Co^4$ performance is compared with the standard Transformer baseline as well as with highly optimized Compact Convolutional Transformer (CCT) both with and without convolutional (Conv) tokenization \cite{hassani2021escaping}. For natural language processing question answering, a synthetic Facebook bAbI dataset \cite{weston2015towards} was used and compared with the standard Transformer. 
\\In experiments, it is observed that the $Co^4$ mechanism achieves superior validation accuracy quickly using only a few attention heads and shallow layers, whereas standard Transformer architectures require a large number of iterations, attention heads, and multiple stacked layers to reach comparable performance levels. 
\\This advantage arises from $Co^4$’s non-linear, bi-directional reasoning, which standard attention mechanisms only approximate after significant depth and width. Notably, scaling the modulatory approach with additional heads and layers yields minimal improvement, suggesting that the core dynamics of $Co^4$ are already highly expressive, even in shallow configurations. 
\\This is because Q–K–V co-adaptation embeds reasoning before the attention operation, reducing the need for iterative layer-wise refinement. In contrast, standard attention can be seen as a sequence of shallow comparisons that gradually build understanding through stacked layers and accumulated context. By comparison, $Co^4$ embeds “deep thinking” at each step, where reasoning is not deferred but baked directly into the triadic interactions, enabling more efficient and coherent inference without excessive architectural complexity.
\subsection{Gradient Flow}
$Co^4$ avoids the need for the second FF residual connection in Transformer model, which is one of the most expensive parts of a Transformer block. All results are presented without the second residual connection. $Co^4$ effectively mitigates unstable gradient flow and reduces spiking gradients, though through a different approach than the gating mechanisms in recurrent neural nets (RNNs) such as Long Short-Term Memory (LSTM) \cite{hochreiter1997long}. 
\\In gated architectures like LSTMs or GRUs, gates control the flow of information over time steps, retaining relevant signals while attenuating irrelevant noise, and prevent vanishing or exploding gradients by modulating updates to the cell state. While $Co^4$ does not use explicit gates, it achieves gradient stability through triadic modulation prior to the attention mechanism. This early modulation filters and reshapes representations, allowing gradients to propagate through contextually meaningful and disentangled pathways, rather than through raw or entangled inputs. $Co^4$ design promotes sparse, context-sensitive activation, reducing unnecessary backpropagation through irrelevant or noisy channels. \\Moreover, the use of the asynchronous MOD function, eq (1), introduces a non-linear, saturating dynamic. It prevents spiking gradients by clipping or modulating abrupt transitions when context and signal are mismatched. Finally, because reasoning is embedded within a single attention layer, $Co^4$ reduces reliance on depth to learn relational meaning. This results in shorter gradient paths and minimizes the risk of signal degradation across multiple layers.
\begin{figure*} 
	\centering
        \includegraphics[trim=0cm 0cm 0cm 0cm, clip=true, width=1\textwidth]{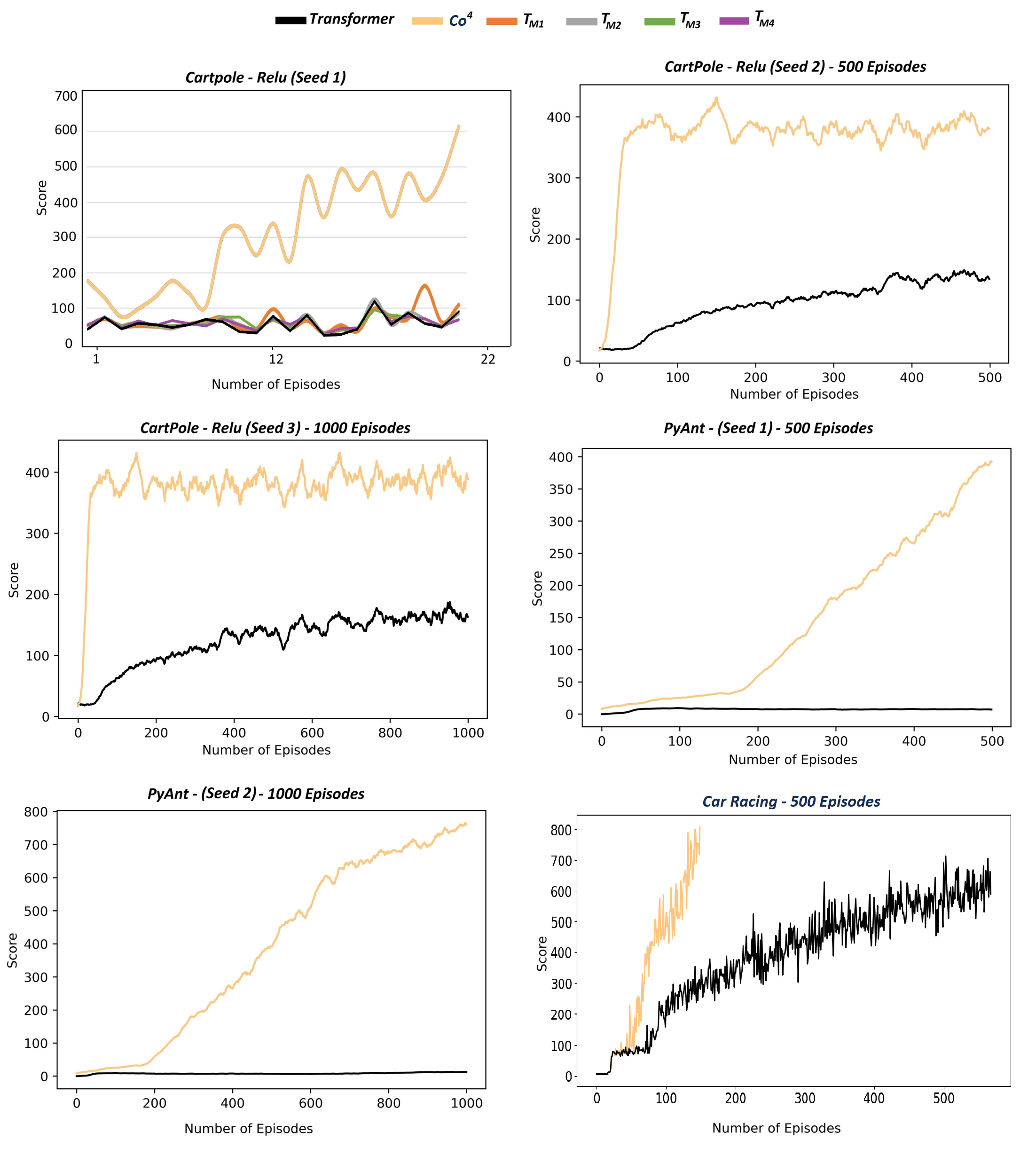}
	\caption{Training Results: In Cart-Pole, PyBullet Ant, and CarRacing (with high dimensional visual environment [96 x 96 x 4]) tasks, $Co^4$, with the same architecture and number of parameters, learns much faster than the Transformer and previously proposed neuro-modulatory functions ($T_{M1}$-$T_{M4}$).}
	\label{l5pc}
 \vspace{-1em}
\end{figure*}
\subsection{The Sensory Neuron as a Transformer: Permutation-Invariant Neural Networks for
Reinforcement Learning}
Here $Co^4$ is applied to permutation-invariant (PI) neural networks (NNs) for RL \cite{tang2021sensory}. In the multisensory RL, V and K are functions of the Sensors 1-N i.e., input $x$ $\epsilon$ $R^N$ (e.g., any linear or non-linear transformations) and Q is the output of latent positional encoding. For permutation invariance, the latent Q is independent of the input $x$ so that permutation $x$ only affects K and V but not Q, which allows the output to be PI \cite{tang2021sensory}. As explained comprehensively in \cite{tang2021sensory}, the individual sensory inputs 1-N or observations $O_t^i$, \textit{i=1, 2, ... N} along with the previous action $a_{t-1}$ passes through a NN module in an arbitrary order such that each NN has partial access to agent's observation at time \textit{t} and $i^{th}$ neuron can only see the $i^{th}$ component of the observation $O_t[i]$, computing $f_K(O_t[i]$, $a_{t-1})$ and $f_V(O_t[i])$ explained in \cite{tang2021sensory}. The overall operation can be described using eq (6-8):
\begin{equation}
K(O_t,a_{t-1})
 =
  \begin{bmatrix}
   f_K(O_t[1], a_{t-1}) \\
   ...\\
 f_K(O_t[N], a_{t-1}) 
   \end{bmatrix} \in  \mathbb{R}^{N \times d_{f_K}} 
\end{equation}
\begin{equation}
V(O_t)
 =
  \begin{bmatrix}
   f_V(O_t[1]) \\
   ...\\
 f_V(O_t[N]) 
   \end{bmatrix} \in  \mathbb{R}^{N \times d_{f_V}} 
\end{equation}
\begin{multline}
    m_t = ReLU6(R(O_t, a_{t-1})^2 + \\ 2R(O_t, a_{t-1}) + C(1+|R(O_t, a_{t-1}))|)
\end{multline}
The architectures of the policy networks, training methods, attention neuron layers, and hyperparameters in all agents are the same as used in \cite{tang2021sensory}. The results presented in Figure 6 were generated using the code provided in \cite{tang2021sensory}, which also serves as the baseline.\\
It was observed that the $Co^4$-driven agent learned the tasks much more quickly than the state-of-the-art Transformer-based PI agents (baseline). Furthermore, the previously proposed context-sensitive neuromodulation transfer functions (2-5) performed comparably to the baseline Transformer model. In the Cart-pole problem, the $Co^4$-Transformer converged to the highest fitness score, exceeding 600 in 22 episodes. In contrast, the baseline learned much more slowly, reaching a fitness score of 100 in 22 episodes and remaining below 200 in 1K episodes. In the PyBullet Ant problem, the $Co^4$-Transformer learned even faster, surpassing 700 in 1K episodes, while the baseline and other $T_{Ms}$ never crossed a fitness score of 100.\\
Testing results for Cart-pole in Table 1 show that the $Co^4$ trained over 1K-10K episodes achieved significantly higher fitness scores with much less standard deviation, both in shuffled and unshuffled scenarios. Although for shuffled inputs, the $Co^4$ performed comparably to the baseline in 1K episodes, it quickly achieved a higher fitness score with less standard deviation in 5K episodes. However, in the PyBullet Ant case (Table 2), the $Co^4$ outperformed in both shuffled and unshuffled scenarios all the time. \\
The car racing task with high dimensional visual environment [96 x 96 x 4] showed similar behavior with different seeds. For example, in fewer than 150 episodes, $Co^4$ reached a score of 800, while the standard Transformer only achieved a score of nearly 600 after 550 episodes (Figure 6 (bottom right)).
\begin{table}
\centering
\caption{CartPole Test Performance (trained over 1K, 5K, and 10K iterations). Each value reports the average score $\pm$ standard deviation across 1K test episodes.}
\resizebox{\columnwidth}{!}{%
\begin{tabular}{lccc}
\toprule
\textbf{Model} & \textbf{1K} & \textbf{5K} & \textbf{10K} \\
\midrule
Transformer & 279$\pm$272 & 340$\pm$308 & 340$\pm$310 \\
Transformer (Shuffled) & 279$\pm$274 & 339$\pm$308 & 340$\pm$309 \\
$Co^4$ & 428$\pm$293 & 524$\pm$408 & 538$\pm$419 \\
$Co^4$ (Shuffled) & 267$\pm$248 & 508$\pm$408 & 536$\pm$417 \\
\bottomrule
\end{tabular}%
}
\end{table}
\begin{table}
\centering
\caption{PyBullet Ant Test Performance (trained over 1K episodes). Each value reports the average score $\pm$ standard deviation across 1K test episodes.}
\resizebox{0.7\columnwidth}{!}{%
\begin{tabular}{lcc}
\toprule
\textbf{Model} & \textbf{ES} & \textbf{ES (Shuffled)} \\
\midrule
Transformer & 121$\pm$53 & 30$\pm$241 \\
$Co^4$ & 1170$\pm$35 & 280$\pm$124 \\
\bottomrule
\end{tabular}%
}
\end{table}
\begin{figure*} 
	\centering
        \includegraphics[trim=0cm 0cm 0cm 0cm, clip=true, width=1\textwidth]{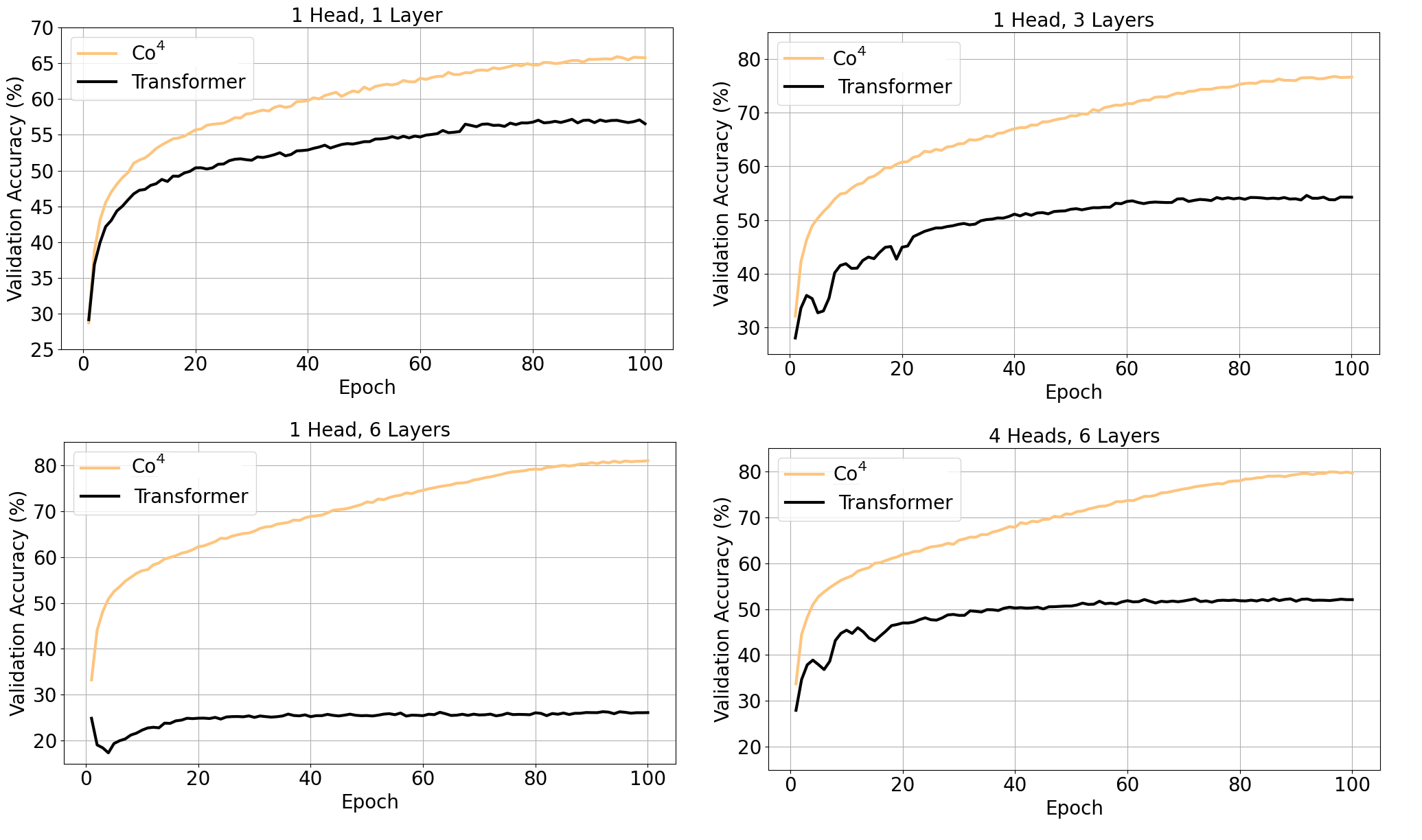}
	\caption{CIFAR-10 Training Results: In all tested configurations, $Co^4$, with the same architecture and number of parameters, learns much faster than the Transformer.}
	\label{l5pc}
 \vspace{-1em}
\end{figure*}
\begin{figure} 
	\centering
	\includegraphics[trim=0cm 0cm 0cm 0cm, clip=true, width=0.25\textwidth]{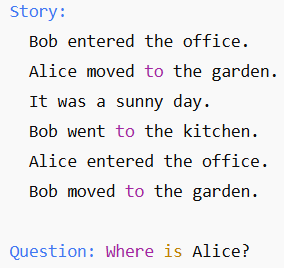}
	\caption{Synthetic dataset similar to Facebook's bAbI dataset \cite{weston2014memory, weston2015towards}.} 
\vspace{-1.4em}
\end{figure}
\begin{figure*} 
	\centering
        \includegraphics[trim=0cm 0cm 0cm 0cm, clip=true, width=1\textwidth]{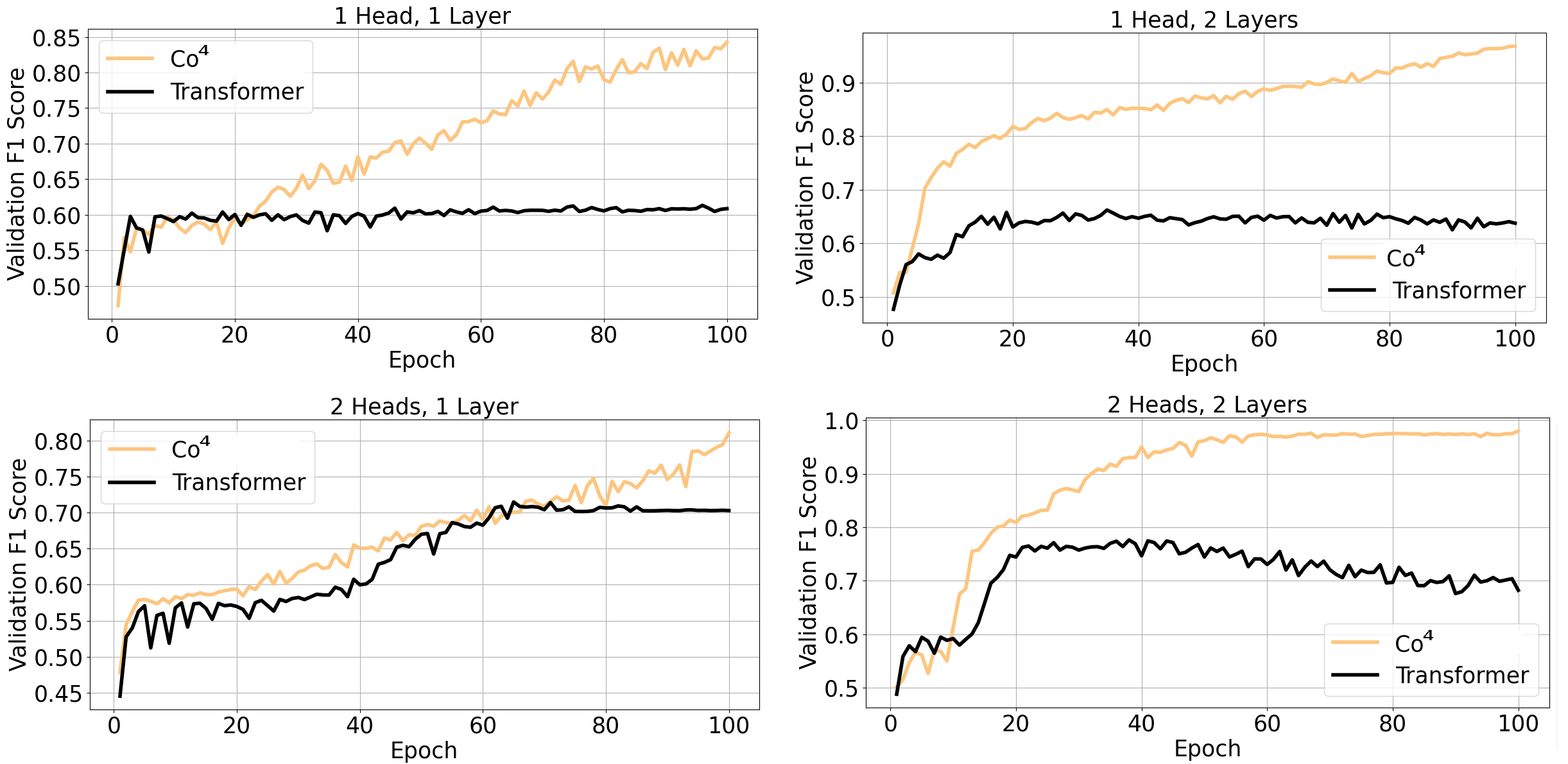}
	\caption{Synthetic Facebook bAbI Training Results: In all tested configurations, $Co^4$, with the same architecture and number of parameters, learns much faster than the Transformer.}
	\label{l5pc}
 \vspace{-1em}
\end{figure*}
\subsection{Image Classification: CIFAR-10}
For CIFAR-10, three transformer-based architectures, standard Transformer, CCT with Convolutional tokenizer \cite{hassani2021escaping}, and $Co^4$-Transformer, were tested. CCT is shown to avoid overfitting and outperform state-of-the-art CNNs on small datasets. It is 10x smaller than other transformers and is 15\% the size of ResNet50 \cite{hassani2021escaping}. \\
The dataset was split into 80\% training and 20\% validation, with a batch size of 64 and a patch size of 4 to generate non-overlapping image patches. Data augmentation techniques were applied to improve generalization. Each model uses an embedding dimension of 256 and varies in terms of the number of transformer layers, latent vectors, and attention heads. The patch embedding module converts each flattened image patch into a learned embedding through a linear transformation, followed by a ReLU activation and dropout to mitigate overfitting. In $Co^4$, learnable latent vectors were introduced. These vectors are repeated to match the batch size and interact with the patch embeddings via attention mechanisms. For each standard transformer layer, separate linear projections were used for the queries, keys, and values. Each layer employs a single layer normalization module (norm1) to support residual connection after attention. Finally, a fully connected layer maps the final output to the number of target classes. For multi-head attention, the query, key, and value tensors are reshaped to accommodate multiple heads, enabling parallel attention computations.\\
For $Co^4$, $K_{mod}$, $Q_{mod}$, and $V_{mod}$ were modified using MOD functions.  Standard attention scores were computed using the dot product of queries and keys, scaled by the head dimension. Softmax was applied to obtain attention weights, and the weighted sum of values was computed. For residual connection and normalization, the attended output was added to the original latent vectors, and layer normalization is applied. For training, evaluation, and inference, the AdamW optimizer was used with weight decay for regularization. The learning rate schedulers Cosine Annealing and ReduceLROnPlateau were employed.
\begin{table}
\centering
\caption{Validation accuracy (\%) on CIFAR-10 (100 epochs) is shown for different numbers of attention heads (H) and layers (L). All Transformer configurations, including CCT, have a complexity of $\mathcal{O}(N^2)$, while $Co^4$ has a complexity of $\mathcal{O}(8 \times N)$, where $N=64$.}
\small
\begin{tabular}{lccccc}
\toprule
\textbf{Model} & \textbf{H} & \textbf{L} & \textbf{Params (M)} & \textbf{Accuracy (\%)} \\
\midrule
Transformer & 1 & 1 & 0.215 & 56 \\
Transformer & 1 & 3 & 0.61  & 52 \\
Transformer & 1 & 6 & 1.20  & 26 \\
Transformer  & 4 & 6 & 1.20  & 54 \\
CCT 7 \cite{hassani2021escaping}  & 4 & 9 & 4.7  & 77 \\
CCT 7-Conv \cite{hassani2021escaping} & 4 & 9 & 4.9  & 82 \\
$Co^4$        & 1 & 1 & 0.215 & 65 \\
$Co^4$        & 1 & 3 & 0.61  & 76 \\
$Co^4$        & 1 & 6 & 1.20  & 81 \\
$Co^4$        & 4 & 6 & 1.20  & 80 \\
\bottomrule
\end{tabular}
\end{table}
\\Figure 7 and Table 3 present a comparative summary of the results for the 100-epochs setup, as part of quick-to-run tests to demonstrate proof of concept. Standard Transformer performance is relatively low, with accuracies between 26\% and 56\%, even with increased parameters. $Co^4$ outperforms the Transformer across all configurations, reaching up to 81\% accuracy with 8 latent tokens and an attention matrix of size (8 x N). $Co^4$ benefits more from increased depth (L) and attention heads (H), suggesting better capacity to learn complex visual patterns better than Transformer. CCT with and without Conv tokenization performs comparably with deeper layers and more heads.
\begin{table}[htb!]
\centering
\caption{Validation accuracy (\%) on the synthetic bAbI dataset (100 epochs), across different numbers of attention heads (H) and layers (L). All Transformer configurations have a complexity of $\mathcal{O}(N^2)$, while $Co^4$ has a complexity of $\mathcal{O}(4 \times N)$, where $N=60$}
\small
\begin{tabular}{lccccc}
\toprule
\textbf{Model} & \textbf{H} & \textbf{L} & \textbf{Params (M)} & \textbf{Accuracy (\%)} \\

\midrule
Transformer & 1 & 1 & 0.19 & 60 \\
Transformer & 1 & 2 & 0.19 & 63 \\
Transformer & 2 & 1 & --   & 70 \\
Transformer & 2 & 2 & 0.38 & 77 (20 ep) \\
$Co^4$        & 1 & 1 & 0.19 & 84 \\
$Co^4$        & 1 & 2 & 0.19 & 96 \\
$Co^4$        & 2 & 1 & --   & 81 \\
$Co^4$        & 2 & 2 & 0.38 & 98 \\
\bottomrule
\end{tabular}
\end{table}
\subsection{Natural Language Processing: Synthetic Facebook bABi dataset}
Figure 8 depicts a synthetic dataset, similar to Facebook's bAbI dataset \cite{weston2014memory, weston2015towards}, used to test $Co^4$ on NLP tasks. Synthetic stories, inspired by the bAbI tasks, were generated to evaluate both reasoning and language understanding. Each example consists of a series of events that can be either location-related (e.g., “Mary moved to the kitchen”) or distractor sentences (unrelated sentences). For each story, one event is designated as the target event: a target person is chosen, and at a random position in the story, an event is generated involving that person moving to a target place. Consequently, a corresponding question is generated (e.g., “Where is Mary?”) with the target place as the answer. \\
The same models used for computer vision were employed except CCT. Basic tokenization and vocabulary construction are applied to process the text. The AdamW optimizer with weight decay for regularization is employed for optimization, and cross-entropy is used as the loss function. A set of random examples is selected to display the original stories, questions, true answers, and the predictions from the trained model. Finally, the overall performance is quantified using the Macro F1 Score.\\
Figure 9 and Table 4 present a comparative summary of the results for the 100-epochs setup. Transformer shows modest performance overall, with accuracies ranging from 60\% to 77\%. Adding layers (L=2) or heads (H=2) does not consistently improve performance for the Transformer. $Co^4$ significantly outperforms the standard Transformer, achieving up to 98\% accuracy, especially when using more attention heads or layers. This suggests that $Co^4$ benefits more from deeper or more complex attention configurations in NLP, indicating better scalability or inductive bias.
\section{Conclusion}

The initial evidence presented here is one of many reasons to believe that emulating the cellular foundations of higher mental states, ranging from high-level perceptual processing to deep, deliberate imaginative reasoning, could be a step toward cognitively meaningful machine intelligence. This approach opens the door not only to implementing large numbers of lightweight, inference-efficient AI modules, but also to moving these systems beyond mere information processing toward contextual reasoning, shifting from raw efficiency to real understanding. This represents a philosophical shift in how models might interpret reality.\\
A fundamental question that warrants further exploration is whether these cellular mechanisms can deepen our understanding of common sense and imaginative reasoning in higher mammals, and whether this can inform the design of future AI systems that are ethical, efficient, and effective. While the development of human-like AI raises important ethical considerations, the evidence presented here suggests that human imagination is deeply rooted in cooperative neural processes, reflecting a democratic, empowering dynamic shaped by evolution \cite{bregman2020humankind, phillips2023cooperative}.\\
Overall, these cellular mechanisms have the potential to advance the field beyond conventional AI toward real understanding, with implications that extend well beyond current expectations. For instance, they may enable us to unlock the computational potential underlying the fundamental principles of the highest forms of conscious life. While humans may or may not attain this level of consciousness and computational capacity due to inherent physical limitations, machines could emulate these processes and help us uncover new dimensions of consciousness, deepening our connection with creation and advancing human progress.\\
Future work will involve extensive evaluation and testing of $Co^4$’s performance on large-scale AI tasks, particularly in the context of Large Language Models (LLMs).

\section{Acknowledgments}
Advanced Research + Invention Agency (ARIA): Nature Computes Better Opportunity seeds. Cooperation is All You Need \cite{adeel2023cooperation} Team. Professor Bill Phillips, Professor Leslie Smith, and Professor Bruce Graham from the University of Stirling. Professor Johan Frederik Storm (Professor in Neurophysiology) from the University of Oslo. Professor Panayiota Poirazi from IMBB-FORTH. Professor Newton Howard from Oxford Computational Neuroscience. Professor John Broome (Emeritus White’s Professor of Moral Philosophy) from Corpus Christi College, University of Oxford. Professor Peter König from the University Osnabrück. Professor Heiko Neumann from Ulm University, and several other eminent scholars for their help and support in several different ways, including reviewing this work, appreciation, and encouragement. I also acknowledge ChatGPT for its assistance with proofreading and coding support.


An early draft of this work was presented at the European Molecular Biology Organization (EMBO) Workshop on Principles of Dendritic Function and Computation, held from 21–24 May 2024.

\textbf{Competing interests}
AA has a provisional patent application for the algorithm used in this paper.


\bibliographystyle{IEEEtran}
\bibliography{NATURE.bib}
\onecolumn 
\pagebreak 

\end{document}